\documentclass[letterpaper, 10 pt, conference]{ieeeconf}  

\IEEEoverridecommandlockouts                              
\usepackage{amsmath} 
\usepackage{amssymb}  
\usepackage{amsbsy}
\usepackage{graphicx}
\usepackage{mathptmx}
 
\usepackage{amsthm}
\usepackage{subcaption}
\newtheoremstyle{exampstyle}
  {.1ex}
  {.1ex}
  {} 
  {} 
  {\bfseries} 
  {.} 
  {.5em} 
  {} 

\theoremstyle{exampstyle}

\title{\bf Interactive Control Approach to 3D Shape Reconstruction\vspace{-1ex}
}

\author{Bipul Islam, Ji Liu, Anthony Yezzi, Romeil Sandhu
\thanks{B. Islam and R. Sandhu are with the Department of Computer Science \& Biomedical Informatics and J. Liu is with the Department of Electrical and Computer Engineering,  Stony Brook University, Stony Brook, NY 11794. A. Yezzi is with the Department of Electrical and Computer Engineering, Georgia Tech, Atlanta GA, 30309. R. Sandhu is also with the Departments of Applied Mathematics \& Statistics, Stony Brook University, Stony Brook, NY 11794. {\tt\small E-mail: bipul.islam@stonybrook.edu}}}

\begin{document}

\maketitle
\begin{abstract}

The ability to accurately reconstruct the 3D facets of a scene is one of the key problems in robotic vision. However, even with recent advances with machine learning, there is no high-fidelity universal 3D reconstruction method for this optimization problem as schemes often cater to specific image modalities and are often biased by scene abnormalities. Simply put, there always remains an ``information'' gap due to the dynamic nature of real-world scenarios.  To this end, we demonstrate a feedback control framework which invokes operator inputs (also prone to errors) in order to augment existing reconstruction schemes.  For proof-of-concept, we choose a classical region-based stereoscopic reconstruction approach and show how an ill-posed model can be augmented with operator input to be much more robust to scene artifacts. We provide necessary conditions for stability via Lyapunov analysis and perhaps more importantly, we show that the stability depends on a notion of absolute curvature.  Mathematically, this aligns with previous work that has shown Ricci curvature as proxy for functional robustness of dynamical networked systems.   We conclude with results that show how our method can improve standalone reconstruction schemes.

\end{abstract}

\section{Introduction}
Sensing the spatial particulars and inferring information about a real-world scene from images is a classical problem in robotic vision with a multitude of uses ranging from motion planning, situational awareness, to medical imaging \cite{yezzi2003stereoscopic,faugeras2002variational,zhao2013overview}.  This said, reconstruction of a complex 3D scene from 2D images is a difficult task due to the amount of uncertainties that must be accounted for in real-world scenarios.  Although much progress have been made over the last few decades, reconstruction methodologies often fail as a result of imaging artifacts including, but not limited to, noise, occlusions, clutter, and non-uniform illumination. In short, no universal algorithm exists which can work seamlessly across all image modalities \cite{zhu2018guiding}. To combat such risk complexities, there is a need for domain experts or an operator who is able to provide an estimate of the ideal result and subsequently able to verify the quality of reconstruction. Here, we aim to ``inject'' 2D operator inputs in-loop to drive a (multi-agent) 3D surface deformation while ensuring the resulting system is stable in the sense of Lyapunov \cite{khalil1996noninear}.  While this work builds off of our previous work in image segmentation \cite{zhu2018guiding} and reconstruction \cite{yezzi2003stereoscopic}, there lies a few tacit yet important discerning caveats.  \textbf{Firstly}, we show that 2D operator inputs of a given set of images can be aptly ``mapped'' to 3D world and such inputs, are stable.  Mathematically, this not a trivial issue as any input on a 2D background should also be corroborated by a 3D action on infinitely large (``blue sky'') background (e.g., specifying the 3D action location based on 2D background input is ill-posed).   From a stability perspective, such singular 2D actions affect not only a 3D surface deformation, but indirectly affect other 2D passive sensors via 3D-to-2D projections during the reconstruction process. \textbf{Secondly}, the control laws are developed in-part based on a notion of absolute principle curvature which is a main underlying theme of this work (e.g., confluence of geometry \& control).  \textbf{Thirdly}, curvature can be shown to relate to a notion of ``trust'' in the sense of how quickly our reconciled solution converges from both the operator and autonomous perspective.  This will be stylized in detail in future work, but is presented here to place this work and contributions in context.  We now briefly revisit a few techniques as it pertains to this work.

\begin{figure}[!t]
\centering
    \includegraphics[width=\linewidth]{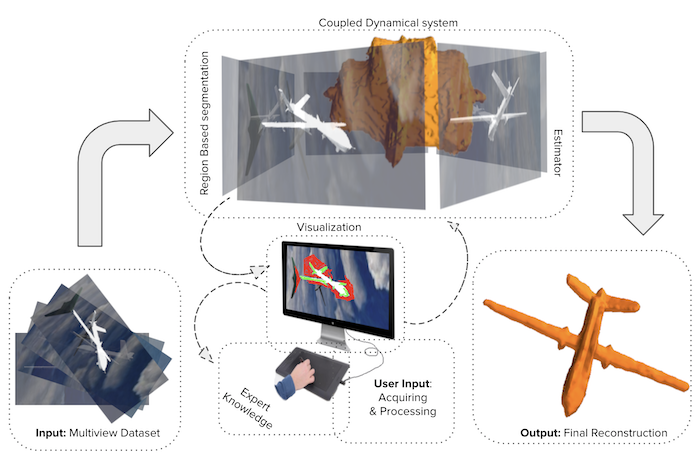} 
	\caption{Schematic outline of interactive feedback control stereoscopic reconstruction framework. }
     \label{fig:sweet} 
\end{figure}
\subsection{Brief 3D Reconstruction Literature Review}
Most modern scene reconstruction methods use the popular deep (reinforcement) learning variants and are often characterized by the requirement of massive training samples \cite{dai2017scannet,zhang2017physically}. Some examples of such systems are ScanNet \cite{dai2017scannet} that uses over 2.5 million scenes to train a system that can understand indoor scenes to \cite{zhang2017physically} where authors furnish a synthetic dataset in order to develop an understanding of surface normal prediction, semantic segmentation, and object boundary detection. Generally, such schemes are highly dependent on the training quality. To combat this, \cite{gwak2017savarse} explores the use of supervision as an alternative for expensive 3D annotation from which perspective projection and back propagation are employed.  On the other hand, such methods use local correspondence matching and hence, are fallible to drawbacks resulting from scene abnormalities (e.g., noise, non-uniform illumination  \cite{chen2015deep}). In regards to robotic vision, such correspondence-based solutions generally involve the well-known concept of SLAM (Simultaneous Localization and Mapping) \cite{aulinas2008slam,zhang2015optimally,zhao2018good}. This said, SLAM-based methods traditionally suffer from the requirement of high computational power for sensing a sizable area and process the resulting data to perform both mapping and localization. Also, there is a tacit requirement that input scene images should have overlap from image-to-image. To this end, SFM (Structure From Motion) based methods provide a relaxed version of this problem \cite{yezzi2003structure,faugeras1993three} (i.e., Google uses this approach in their popular street-view application on Google maps \cite{klingner2013street}).  More recently, \cite{choy20163d} explores a recurrent neural network (3D-R2N2) by employing shape priors in which one learns 2D to 3D mapping from images of objects to their underlying 3D shapes from large collections of synthetic data.  In particular, the authors have been seemingly able to show their method outperforming SLAM or SFM (albiet with learnt knowledge) when there is lack of texture or baseline. 

Nevertheless, this paper does not argue the rigors of the underlying reconstruction method itself and our particular focus on our previous work \cite{yezzi2003stereoscopic} is in-part due a correspondence-free method, independence to local (image-gradient) structure, and dependence on geometric techniques connected to image segmentation \cite{mumford1989optimal,chan1999active,bertalmio2001variational}. Undoubtedly, each approach whether it be SLAM-based, deep (reinforcement) learning variants, and/or geometric methods work optimally with respect to the prospective operating environment (e.g., space, low-power requirements compared ground-based robotic vision). At the same time, any such reconstruction are not infallible to errors that arise in real-world dynamic scenes from a human-perception standpoint.  This said, human-perception is also fallible and any operator input based on a visual estimate is prone to errors. Philosophically, we make the argument that terms such as over-fitting and uncertainty are in part, perceived by an expert who generally acts as a passive entity in such methods.  Thus, the problem we seek to resolve is to not only rectify the expected and ideal reconstruction in real-time \cite{nguyen2012robust}, but provide the necessary feedback control characterization when invoking operator input \cite{doyle2013feedback}.

The remainder of the paper is organized as follows:  In the next, we introduce stereoscopic reconstruction via classic image segmentation.  Then Section III provides a control framework along with the necessary conditions for stability.   Section IV  presents experimental results.  From this, we conclude with future work in Section V.

\section{From Segmentation to 3D Reconstruction}
This section presents a general introduction to geometric stereoscopic segmentation.
\subsection{Geometric 2D Image Segmentation}
Let us begin with the classic binary problem of segmenting an image $I:\Omega\mapsto\mathbb{R}^n$ into a foreground and background described by functionals $r_{o}:\zeta$, $\Omega\mapsto\mathbb{R}$ and $r_{b}:\zeta$, $\Omega\mapsto\mathbb{R}$ which measure the similarity of of the image pixels  with a statistical model over the regions $R$ and $R^{c}$, respectively. Here,  $\zeta$ corresponds to the photometric variable of interest. Then, one can define a partitioning problem where the optimal partition between foreground/background is described by a partial differential equation \cite{bertalmio2001variational,kichenassamy1996conformal}; i.e.,
\begin{align}
\label{eq:Segmentation_Energy}
E &= \int_{R}r_{o}(I(\mathbf{x}),C) + \int_{R^{c}}r_{b}(I(\mathbf{x}),C)d\Omega  \\
\frac{\partial{E}}{\partial C} &= \beta \vec{N} \nonumber
\end{align}
where $\beta: \mathbb{R}^2\mapsto\mathbb{R}$ can be considered ``forces'' along the curve (partition boundary) that describe the direction of the corresponding evolution in the normal $\vec{N}$ direction.  While a complete review of such methodology is beyond the scope of this note, we do refer the reader to several seminal references \cite{mumford1989optimal,chan1999active}. For the case image segmentation, it suffices to understand that the partitioning curve $C$ ``lives'' in the 2D image domain.
\subsection{Stereoscopic 3D Reconstruction}
Now, if we consider the problem of 3D reconstruction from 2D images, one can redefine the functional in equation (\ref{eq:Segmentation_Energy}) as follows:
\begin{equation}
\label{eq:Gradient_Energy}
E = \sum_{i=0}^N\int_{R_i}r_{o}(I_i(\mathbf{\hat{x}_i}),\pi_i^{-1}(\mathbf{\hat{x}_i}),\hat{c}_i) + \int_{R_i^{c}}r_{b}(I_i(\mathbf{\hat{x}_i}),\Theta_i(\mathbf{\hat{x}_i}),\hat{c}_i)d\Omega_i
\end{equation}
where the difference is the functional now depends on $N$ image observations $I_i$ and where a particular 2D image silhouette curve $\hat{c_i}$ is derived from a single 3D occluding curve $C$ (with a slight abuse of notion) on a given smooth surface $S$ in $\mathbb{R}^{3}$ with a corresponding 3D background $B$ treated as infinitely large sphere with angular coordinates $\Theta =  (\gamma,\upsilon)$.  That is, $\hat{c_i} = \pi_i(C)$ where $\pi_i:\;\mathbb{R}^3 \mapsto \Omega_i$ is the realization of the $i$-th pin-hole camera (sensor) that projects the 3D world onto the 2D domain.  Similarly, the background can be related in a one-to-one manner with the image coordinates $\mathbf{\hat{x}_i}$ of each observation through the mapping $\Theta_i$ (``blue sky'' assumption).  To be more precise,  $\mathbf{x} = (x,y,z)$ is surface coordinates of $S$ in $\mathbb{R}^3$ and further note that $\mathbf{x_i} = (x_i,y_i,z_i)$ denote the same points expressed in $i$-th calibrated camera coordinates relative to the $i$-th image. Moreover, $\mathbf{\hat{x_i}}=(\hat{x_i},\ \hat{y_i})=(x_i/z_i, y_i/z_i$)  is the aforementioned perspective projection due to the $i$-th pin-hole camera $\pi_i$.  In turn, $r_{o}$ and $r_{b}$ redefined to be radiance functions.  That is, the foreground object of interest supports a radiance function of $r_{o}$: $S \rightarrow \mathbb{R}$ with the usual area element $dA$.  Similarly, the background supports a different radiance function $r_{b}$: $B \rightarrow \mathbb{R}$.  As such, for a given 3D surface, it is possible to partition each image domain $\Omega_i$ of $I_i$ into a foreground object region $R_i = \pi_i (S) \subseteq \Omega_i$ and the corresponding background region $R_i^c$. Note, the operator $\pi_i$ is not one-to-one and, hence non-invertible. However, we can define a back projection operator $\pi_i^{-1}$ using the back tracing of rays from image to the surface, i.e, we have $\pi_i^{-1}: R_i \rightarrow S$ which is a pseudo one-to-one operation.  

Putting this together, assuming the calibrated cameras, the deformation of the surface towards a reconstructed shape based on a set of $N$ image observations can be shown to be of the following form:
\begin{figure}[!t]
\centering
    \includegraphics[width=\linewidth]{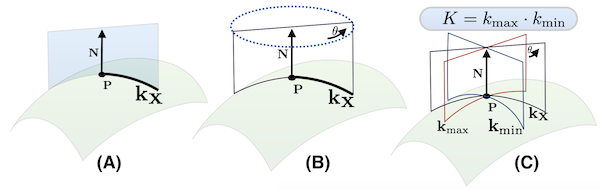} 
	\caption{Visualization of Normal, Principle, and Gaussian Curvature.  (A) Given Point $P$ and Normal $N$, define a perpendicular plane intersection at point $P$.  The curve that plane intersects on the manifold is known as the normal curvature $\kappa_{X}$ in the direction $X$.  (B) We can define other normal curvatures through a rotation of the plane by $\theta$.  (C) The max and min normal curvature are what is known as principle curvatures.  The product of these principle curvatures yields Gaussian curvature. }
     \label{fig:curvature} 
\end{figure}
\begin{equation}
\label{eq:Gradient_Surface}
\frac{\partial S}{\partial t} = \sum_{i=0}^N \beta_i \cdot \bigg( \frac{\nabla_{x_i} \chi_i \cdot {\mathbf{x_i}}}{z_i^3}\bigg) \vec{N} \nonumber
\end{equation}
where we define a visibility characteristic function $\chi_i$ from a given location $\mathbf{x_i}$ on a surface $S$ as:
 \begin{equation}\label{eq:chi}
\chi_i(\pmb x) = \begin{cases} 
1, & \mbox{if } \mathbf{x_i} \in \pi_i^{-1}(R_i) \\ 
0, & \mbox{if } \mathbf{x_i} \notin \pi_i^{-1}(R_i). \nonumber
\end{cases} 
\end{equation}
This can be re-written in terms of the smooth regularized-Heaviside function $H$ along with (outward) surface normals $\vec{N}$ at each point $\mathbf{x_i}$ of the surface $S$:
\begin{equation}\label{eq:xi}
\chi_i = 1 - H({\mathbf{x_i}}\cdot \vec{N}) \nonumber
\end{equation}
Given the above, we are now able to formulate a control-based reconstruction scheme from which a given physical 2D action, based on visual perception (information), can be used to interactively ``sculpt'' a 3D shape in collaboration with the above autonomous 3D reconstruction algorithm.
\section{Control-Based Reconstruction}
Let us begin by redefining the general form of a surface reconstruction evolution above in level-set notation as follows: 
\begin{equation}\label{eq:src}
\frac{d\phi}{dt} = \sum_{i=0}^N \psi_i (\mathbf{\hat{x_i}},\mathbf{x_i},t)  \delta(\phi(\mathbf{x}))
\end{equation}
where $\psi_i: \mathbb{R}^{3}\rightarrow \mathbb{R}$ is the surface gradient information computed from the photometric image data, $\phi: \mathbb{R}^{3}\rightarrow \mathbb{R}$ is a level-set function, and $\delta(.)$ is the classical Kronecker delta function.
Hence, to ``close the loop'' that incorporates a physical 3D operator performing 2D inputs in order to control the 3D evolution dynamics of the evolving surface, one has
\begin{equation}\label{eq:ctrl_dyn}
\frac{d\phi}{dt} = \sum_{i=0}^N [ \psi_i + F_i(\phi, \phi^\ast)] \delta(\phi)
\end{equation}
where $F_i$ is the to be defined control law that drives $\phi$ towards the ideal (perfect) surface $\phi^\ast$ as $t\rightarrow \infty$. The definition of an ideal surface is this note is a result with no errors. For this work, we use the mean-separable segmentation energy \cite{chan1999active} as our reconstruction model. From this, $\nabla_{x_i} \chi_i \cdot {\mathbf{x_i}}$ can be expressed in terms of curvature for points on the surface which leads us to the following Lemma.
\\

\noindent\textbf{Lemma III.1} \emph{For a given characteristic function $\chi_i$ and a point $\mathbf{x_i}\in S$ that lies on the corresponding surface ``imaged'' from a given camera $\pi_i$, we have that}
\begin{equation}
\nabla_{x_i }\chi_i \cdot \mathbf{x_i} = - \kappa_u \left\lVert \mathbf{x_i}\right\rVert^2 \delta(\mathbf{x_i} \cdot \vec{N}).
\end{equation}

\noindent\emph{Proof:} Following the nomenclature defined above and noting $\mbox{II}({\pmb x},{\pmb x})$ is the second fundamental form \cite{faugeras1993three,do2016differential}, we have
\begin{equation}
\begin{split}
\nabla_{x_i} \chi_i \cdot \mathbf{x_i} &= \langle \nabla_{x_i}(1 -  H(\mathbf{x_i}\cdot \vec{N})),\mathbf{x_i} \rangle\\
&=-\langle \delta(\mathbf{x_i}\cdot \vec{N})\nabla_{x_i} (\mathbf{x_i} \cdot \vec{N}) ,\mathbf{x_i} \rangle\\
&=-\delta(\mathbf{x_i} \cdot \vec{N})\langle\nabla_{x_i} (\mathbf{x_i} \cdot \vec{N}) ,\mathbf{x_i} \rangle\\
&=-\delta(\mathbf{x_i} \cdot \vec{N}) (\nabla_{x_i} \vec{N}^T \mathbf{x_i})^T \mathbf{x_i}\\
&=-\delta(\mathbf{x_i} \cdot \vec{N}) [ \mathbf{x_i}^T \nabla_x \vec{N} \mathbf{x_i}]\\
&= -\delta(\mathbf{x_i} \cdot \vec{N}) \begin{bmatrix}u & v \end{bmatrix} \begin{bmatrix}l & m\\ m & n\end{bmatrix} \begin{bmatrix}u \\ v \end{bmatrix}\\
&= -\delta(\mathbf{x_i} \cdot \vec{N})\  \mbox{II}(\mathbf{x_i},\mathbf{x_i})\\
&= -\delta(\mathbf{x_i} \cdot \vec{N}) \frac{ \mbox{II}(\mathbf{x_i},\mathbf{x_i})}{\mathbf{x_i}^T \mathbf{x_i}}\left\lVert \mathbf{x_i}\right\rVert^2\\
&= - \delta(\mathbf{x_i} \cdot \vec{N}) \kappa_u \left\lVert \mathbf{x_i}\right\rVert^2
\end{split}
\end{equation}

\noindent where $k_u$ is the \textbf{normal curvature} \emph{in a particular viewing direction $\mathbf{x_i}$} on the corresponding surface $S$.  We refer to Figure \ref{fig:curvature} for a visualization of this type of curvature on a given manifold. From this, we can rewrite $\psi_i$ as the following:
\begin{equation}\label{eq:yez1}
\psi_i = - \beta_i\frac{  \delta(\mathbf{x_i}\cdot \vec {N}) \kappa_u \left\lVert \mathbf{x_i}\right\rVert^2}{z_i^3}.
\end{equation}
Furthermore, as we aim to define a control law $F_i$ such that $\lim_{t \rightarrow \infty} \phi({\pmb x})\rightarrow \phi^*({\pmb x})$, we define the error between our current estimate and ideal shape (no errors) as
\begin{equation}
E_e(\mathbf{x},t) := H(\phi(\mathbf{x},t)) - H(\phi^\ast({\mathbf{x}})).
\end{equation}

\noindent In doing so, we are now able to define the existence of the control law $F_i$ via Lyapunov method of stabilization.
\\

\noindent\textbf{Theorem III.1:} \emph{Let us assume $z_i\geq 1$ and $||\mathbf{x_i}||^2 \leq z_i^3$ as well as let $\kappa_{max}$ and $\kappa_{min}$ be the the \textbf{principle maximum curvature} and \textbf{principle minimum curvature} at a given point $\mathbf{x_i}$ with respect to an imaging referential camera $\pi_i$, respectively.  Then the control law}
\begin{equation}\label{eq:ctrl_law}
F_i = -\mid \beta_i \mid \kappa_{abs} E_e
\end{equation}
\emph{where $\kappa_{abs} = \mid\kappa_{min}\mid + \mid\kappa_{max}\mid$, asymptotically stabilizes the system given in equation (\ref{eq:ctrl_dyn}) from the current evolving surface $\phi({\pmb x},t)$ to the ideal surface, $\phi^\ast({\pmb x})$ as $t\rightarrow \infty$}.
\\

\noindent\emph{Proof: }
We choose the Lyapunov function $V(E_e, t) \in C^1$ defined in terms of $E_e(\mathbf{x},t)$ as
\begin{equation}
V = \frac{1}{2} \int_{S \cup S^\ast} \left\lVert E_e(\mathbf{x},t) \right\rVert^2 dx.\\
\end{equation}

\noindent Differentiating $V$ with respect to time $t$ we get:
\begin{equation}
\begin{split}
\frac{\partial V}{\partial t} &= \int_{S \cup S^\ast} E_e \frac{\partial E_e}{\partial t }dx\\
&= \int_{S} E_e [\delta(\phi) \frac{\partial \phi}{\partial t }] dx\\
&= \int_{S} E_e \delta(\phi)[\sum_{i=0}^N [ \psi_i + F_i] \delta(\phi) ] dx\\
\end{split}
\end{equation}
The simplification over the union $S \cup S^\ast$  results from the application of the Kronecker delta function.  Moreover, one can show that resulting system is stable (i.e., $V$ has a negative semidefinite derivative):
\begin{equation*}
\begin{split}
\frac{\partial V}{\partial t}  &= \sum_{i=0}^N \int_{S}E_e \delta(\phi)^2  [ \psi_i + F_i] dA\\
&= \sum_{i=0}^N \int_{S}E_e \delta(\phi)^2  \bigg[  \beta_i \cdot \frac{\nabla_{x_i} \chi_i \cdot {\mathbf{x_i}}}{z_i^3} + F_i\bigg] dA\\
&= \sum_{i=0}^N \int_{S}E_e \delta(\phi)^2  \bigg[  \beta_i \cdot \frac{\nabla_{x_i} \chi_i \cdot {\mathbf{x_i}}}{z_i^3}  -\mid \beta_i \mid \kappa_{abs} E_e\bigg] dA\\
&= \sum_{i=0}^N \int_{S}\delta(\phi)^2  \bigg[ E_e \cdot \beta_i  \cdot \frac{\nabla_{x_i} \chi_i \cdot {\mathbf{x_i}}}{z_i^3}  -\mid \beta_i \mid \kappa_{abs} E_e^2\bigg] dA\\
& \leq \sum_{i=0}^N \int_{S}\delta(\phi)^2  \bigg[E_e^2 \cdot \mid \beta_i\mid \biggl\vert \frac{\nabla_{x_i} \chi_i \cdot {\mathbf{x_i}}}{z_i^3}\biggr\vert  -\mid \beta_i \mid \kappa_{abs} E_e^2\bigg] dA\\
&=  \sum_{i=0}^N \int_{S}\delta(\phi)^2  E_e^2  \mid \beta_i\mid \bigg[ \biggl\vert \frac{\nabla_{x_i} \chi_i \cdot {\mathbf{x_i}}}{z_i^3}\biggr\vert  -\kappa_{abs}\bigg] dA\\
&=  \sum_{i=0}^N \int_{S}\delta(\phi)^2  E_e^2  \mid \beta_i\mid \bigg[  \mid \kappa_u  \frac{\left\lVert\mathbf{x_i}\right\rVert^2}{z^3} \mid  -\kappa_{abs}\bigg] dA\\
& < \sum_{i=0}^N \int_{S}\delta(\phi)^2  E_e^2  \mid \beta_i\mid \big[  \mid \kappa_u   \mid  -\kappa_{abs}\big] dA\\
&\leq 0
\end{split}
\end{equation*}

\noindent
In particular, the above control law will be dependent on curvature.  While beyond the scope of this note, one can show exponential convergence whereby higher curvature coincides with faster convergence rates.  While we have not included this derivation in the present work due to scope and for sake of clarity, we will expound upon this in future work.  This said, we present such comments to better highlight important caveats in terms of geometry and control as well as how one can start to define notions of ``trust'' (from a reconciliation of an operator augmentation) to that of a geometric (curvature) quantity.  We would like to highlight there exists analogous behavior in networked dynamical systems in which one is able to use discrete Ricci curvature as a measure for network robustness \cite{sandhu2015graph}.  In such work, one can leverage the concept of k-convexity similarly to above to define positive correlation between Boltzmann entropy, curvature, and rate functions from thermodynamics.  Ultimately, this work will seek to build upon this area and in particular, explore notions of ``trust'' in the sense of geometric quantities such as curvature.  

Nevertheless, in designing operator guided inputs, we note perfect knowledge of ideal surface is not readily available (even from a human visualization perspective) due a myriad of reasons including, but not limited to, occlusions, clutter, and/or inability to define a well-posed model across image modalities.  As such, we allow an operator (whom is also prone to errors) to make interactions with the system in order to reconcile \emph{one's belief} with built autonomy towards an estimate of the ideal surface.  We stress the fact that the input from a human is fallible and such input indirectly affects our control law through the adjudication of an ``ideal'' estimate.  This estimate herein is denoted as $ \hat\phi^\ast(\mathbf{x},t)$. Moreover, we define $\epsilon_i^k(\mathbf{\hat{x_i}},t)$ as the $k$-th input on a given image $i$ and the accumulated input $U_i:\mathbb{R}^2\rightarrow\mathbb{R} $ as 
\begin{align}
\epsilon_i^k(\mathbf{\hat{x_i}},t) & := \pm p \text{  (constant)} \nonumber \\
U_i(\mathbf{\hat{x_i}},t) &:= \sum_{l=0}^{k} \epsilon_t^k(\mathbf{\hat{x_i}}). \nonumber 
\end{align}

That is, we seek to allow for \emph{the physical operator to make 2D actions such that it will deform a 3D surface}.  In other words, we are able to define a 3D control law based on 2D inputs which is particularly helpful as the operator is generally ill-equipped to alter the 3D shape itself (i.e., we assume the operator not to be an artist).  To derive the coupled system that fuses 2D operator input to control the 3D surface deformation, we must also define the errors for \textbf{both} the operator and autonomous model:
\begin{equation}\label{eq:err}
    \begin{split}
        E_A &= H(\phi(\mathbf{x}),t)) - H(\hat{\phi}^{\ast}(\mathbf{x}),t)) \\
        E_{u_i} &= H(\hat{\phi}^{\ast}({\pmb x}),t)) - H(U_i(\pi_i({\pmb x}),t)).\\
    \end{split}
\end{equation}
\begin{figure}[!t]
\center
\begin{subfigure}{0.30 \linewidth}
    \includegraphics[width=\linewidth]{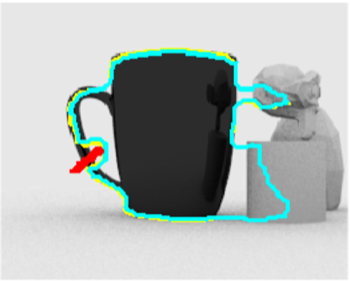} 
   \caption{Incision} 
    \label{fig:0}
   \end{subfigure} 
    \begin{subfigure}{0.30 \linewidth}
    \includegraphics[width=\linewidth]{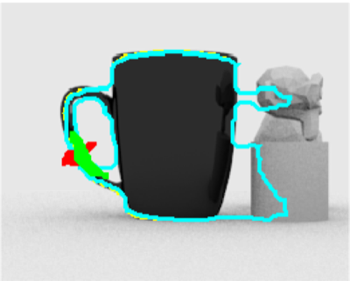} 
   \caption{Repair} 
    \label{fig:1} 
  \end{subfigure}

  \begin{subfigure}{0.30\linewidth}
    \includegraphics[width=\linewidth]{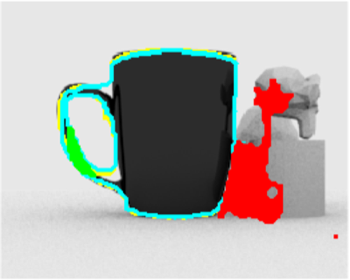} 
  \caption{Consolidate}
    \label{fig:2} 
  \end{subfigure} 
  \begin{subfigure}{0.30\linewidth}
    \includegraphics[width=\linewidth]{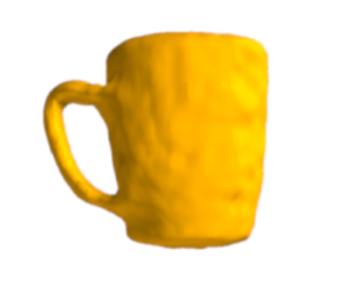}  
     \caption{Final} 
    \label{figinit:d} 
   \end{subfigure} 
   \caption{A summary of operators actions to maneuver out of the local minima in a complex occluded scene. The images (A), (B), and (C) are views of the model after each of interaction milestones. Sub-figure (C) shows the final reconstruction. }
  \label{fig:teacup1} 
\end{figure}
Given this, we can now define a coupled PDE system that unifies both the operator based inputs along with that of the autonomous counterpart which is representative of an estimator-observer behavior as follows: 
\begin{subequations}
\begin{align}
\frac{\partial\phi}{\partial t} &= \sum_{i=1}^N [\psi_i + F_i] \delta(\phi) \label{eq:coupled1}\\
\phi({\boldsymbol x},0) &= \phi^0({\boldsymbol x}) \nonumber \\
\frac{\partial \hat\phi^\ast}{\partial t} &= \sum_{i=1}^N [E_A + f_i(U_i, E_{u_i})]\label{eq:coupled2}\\
\hat{\phi^\ast}({\boldsymbol x},0) &= \phi^0({\boldsymbol x}) \nonumber
\end{align}
\end{subequations}
where the tuning function $f_i(U_i, E_{u_i})$ that is dependent on operator input from an image observation can be defined as 
\begin{equation}\label{eq:tune}
f_i(U_i, E_{u_i}) = - \mid U_i\mid E_{u_i}.
\end{equation}
This said, the above system then needs to be shown that it is is still stable even from imperfect operator actions.  To do so, we define the accumulated total errors for both the operator and autonomous model as

\begin{align}\label{eq:erei}
E(t) &:= \frac{1}{2} \sum_{i=1}^N \int_{S \cup B}\mid U_i\mid E_{u_i}^2 dx \\
\Gamma(t) &:= \frac{1}{2} \sum_{i=1}^N \int_{S \cup S^\ast}E_A^2 .
\end{align}
From this, we now arrive at the following result.
\\

\noindent\textbf{Theorem III.2:} \emph{Let us assume previous notation and results in Theorem III.1 and further assume that operator input has stopped (i.e., $U_i$ is constant in all viewing directions), then the estimator }
\begin{align}\label{eq:coupled}
\frac{\partial \hat\phi^\ast}{\partial t} &= \sum_{i=1}^N [E_A + f_i(U_i, E_{u_i})] \nonumber
\end{align}
\emph{where $f_i(U_i, E_{u_i}) = - \mid U_i\mid E_{u_i}$ will stabilize the resulting coupled system in equation (\ref{eq:coupled1}) and equation (\ref{eq:coupled2}).  Namely, the total error $\Phi(t):= E(t) + \Gamma(t)$ has a negative semidefinite derivative.}

\begin{proof}\vspace{-.2ex}
Let us begin by differentiating $E(t)$ with respect to $t$:
\begin{equation}\label{eq:erei_1}
\begin{split}
\frac{\partial E}{\partial t}&= \sum_{i=1}^N \int_{S\cup \hat{S}^\ast}\mid U_i\mid E_{u_i} \frac{\partial E_{u_i}}{\partial t}\\
& = \sum_{i=1}^N  \int_{S\cup \hat{S}^\ast} \mid U_i\mid E_{u_i} \delta(\hat{\phi}^\ast)\frac{\partial \hat{\phi}^\ast}{\partial t} dx\\
& = \sum_{i=1}^N  \int_{S\cup \hat{S}^\ast} \mid U_i\mid E_{u_i} \delta(\hat{\phi}^\ast)[ E_A - \mid U_i \mid E_{u_i}] dx.
\end{split}
\end{equation}
Similarly, differentiating $\Gamma(t)$ with respect to $t$:
\begin{equation}\label{eq:erev_1}
\begin{split}
\frac{\partial \Gamma}{\partial t} &= \sum_{i=1}^N \int_{S\cup \hat{S}^\ast} E_A \bigg[ \delta(\phi)\frac{\partial \phi}{\partial t} -  \delta(\hat{\phi}^\ast)\frac{\partial \hat{\phi}^\ast}{\partial t}\bigg]dx\\
&= \sum_{i=1}^N \int_{S\cup \hat{S}^\ast}\delta(\phi)^2 E_A[\psi_i + F_i] dx\\ 
&- \sum_{i=1}^N \int_{S\cup \hat{S}^\ast} \delta(\hat{\phi}^\ast)^2 E_A[E_A - \mid U_i \mid E_{u_i}]dx.
\end{split}
\end{equation}
\begin{figure}[!t]
\centering
    \includegraphics[width=.8\linewidth]{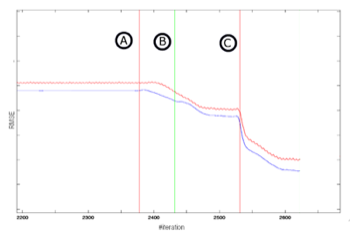} 
   \caption{The overall sequence w.r.t. to energy minimization of operator action corresponding to Figure \ref{fig:1}.  (A ) Incision, (B) Repair, (C) Consolidate.}
  \label{fig:energyLandscape} 
\end{figure}
\begin{figure*}[!t]
\centering
      \begin{subfigure}{0.14\linewidth}
      \centering
      \includegraphics[width=\linewidth]{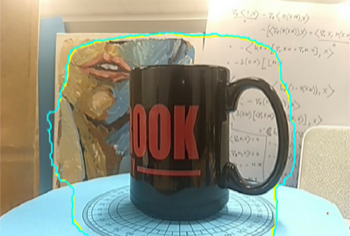}
       \end{subfigure}
      \begin{subfigure}{0.14 \linewidth} 
    \centering
    \includegraphics[width=\linewidth]{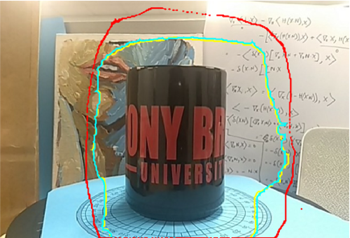} 
   \end{subfigure} 
    \begin{subfigure}{0.14\linewidth}
    \centering
    \includegraphics[width=\linewidth]{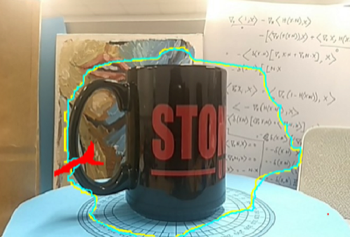} 
  \end{subfigure}
  \begin{subfigure}{0.14\linewidth}
    \centering
    \includegraphics[width=\linewidth]{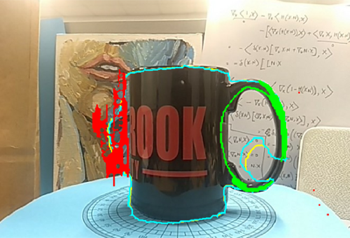} 
  \end{subfigure} 
  \begin{subfigure}{0.14\linewidth}
   \centering
    \includegraphics[width=\linewidth]{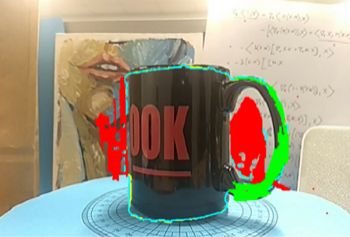} 
   \end{subfigure}  
\begin{subfigure}{0.14\linewidth}
    \centering
    \includegraphics[width=\linewidth]{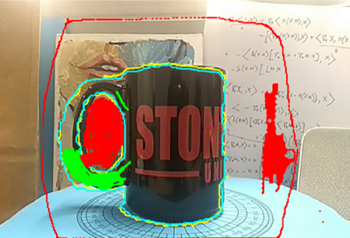} 
  \end{subfigure} 
  
    \begin{subfigure}{0.14 \linewidth}
    \centering
    \includegraphics[width=\linewidth]{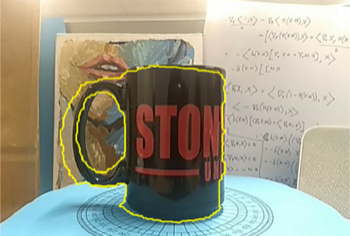} 
  \end{subfigure}
  \begin{subfigure}{0.14\linewidth}
\centering
    \includegraphics[width=\linewidth]{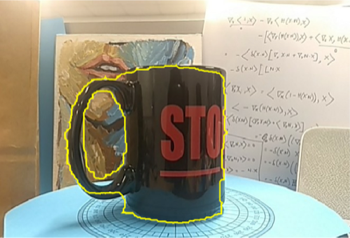} 
  \end{subfigure} 
    \begin{subfigure}{0.14\linewidth}
\centering
    \includegraphics[width=\linewidth]{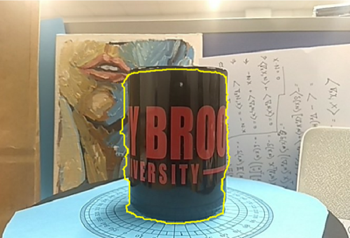} 
   \end{subfigure} 
  \begin{subfigure}{0.14\linewidth}
 \centering
    \includegraphics[width=\linewidth]{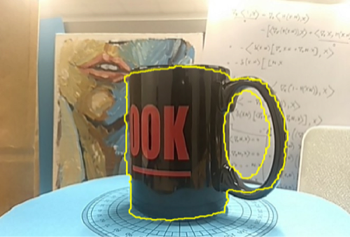} 
   \end{subfigure} 
     \begin{subfigure}{0.14\linewidth}
\centering
    \includegraphics[width=\linewidth]{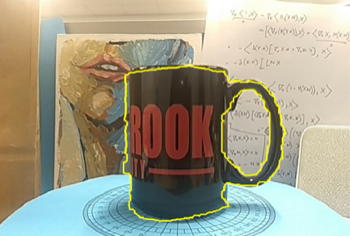} 
   \end{subfigure} 
    \begin{subfigure}{0.14\linewidth}
 \centering
    \includegraphics[width=\linewidth]{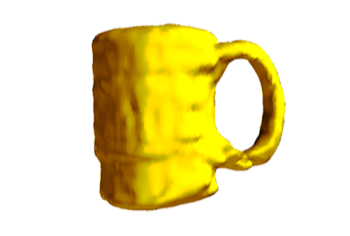} 
  \end{subfigure}

   \caption{3D reconstruction of a cup in clutter and camera miscalibration. Top row: Sequence of user initiated operations to reorient the flow at multiple time instances. Bottom row: Final silhouette curves and reconstruction. Note:  Yellow Curve is Autonomous Surface, Blue Curve is Ideal Estimate, Green is Foreground Interaction, Red is Background Interaction.}
  \label{fig:real_cup} 
\end{figure*}

\noindent From this, we are now able to combine terms for the total labeling error $\Phi(t) = E(t) + \Gamma(t)$.  That is, summing equation \eqref{eq:erei_1} and equation \eqref{eq:erev_1} and simplifying:
\begin{equation}
\begin{split}
\frac{\partial \Phi}{\partial t} & =  \sum_{i=1}^N  \int_{S\cup \hat{S}^\ast} \mid U_i\mid E_{u_i} \delta(\hat{\phi}^\ast)[ E_A - \mid U_i \mid E_{u_i}] dx\\
&+  \sum_{i=1}^N \int_{S\cup \hat{S}^\ast}\delta(\phi)^2 E_A[\psi_i + F_i] dx\\
&-  \sum_{i=1}^N \int_{S\cup \hat{S}^\ast} \delta(\hat{\phi}^\ast)^2 E_A[E_A - \mid U_i \mid E_{u_i}]dx\\
&\leq -  \sum_{i=1}^N \int_{S\cup \hat{S}^\ast} \delta(\hat{\phi}^\ast)^2[E_A - \mid U_i \mid E_{u_i}]^2dx\\
& \leq 0
\end{split}
\end{equation}
As the derivative is negative semidefinite, the coupled system defined above is stable.  
\end{proof}


\addtolength{\textheight}{-3cm}   

\section{Experimental Results and Discussion}
In this section, we demonstrate the proposed algorithm on a variety of scenarios.  In all demonstrated results, green patches, or marks, are made by the user to denote regions in the foreground. Similarly, red denotes regions on images that are to be considered a part of the background. In images where silhouettes are displayed, the yellow silhouette denotes the autonomous surface while the estimate of ideal surface is always presented in cyan.   Each reconstruction utilizes $N=36$ images with the resulting MATLAB code run on an iMac 4.2 Ghz Core i7 with 32GB memory. 

We begin with an example that highlights the method in face of \textbf{occlusions} by objects obfuscating several different imaging views.  This can be seen Figure \ref{fig:teacup1} along with how such inputs affect the energy minimization landscape in Figure \ref{fig:energyLandscape}.  Here, naive reconstruction fails due to ambient occlusion whose intensity is similar to the background.  While there exists varying approaches and shape prior models to overcome such a problem, defining such models for particular scenarios becomes quite cumbersome and yet, may not yield stable results.  We are able to properly reconstruct the shape through operator input with a simplified model as defined in \cite{chan1999active}.  For this experiment, the user made $12$ interactions for the foreground and $47$ interactions for the background. In particular, in regards to the operator input and its impact on the energy landscape, the user actions can be partitioned into 3 milestones: initial incision (Figure: \ref{fig:teacup1}\subref{fig:0}), followed by a repair of the surface (Figure: \ref{fig:teacup1}\subref{fig:1}), and then, consolidating the surface by helping it ``free'' itself from scene anomalies (Figure: \ref{fig:teacup1}\subref{fig:2}).  

\begin{figure}[!t]
\center
\begin{subfigure}{0.30 \linewidth}
    \includegraphics[width=\linewidth]{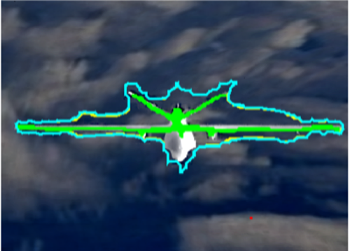} 
    \label{figinit:a0}
   \end{subfigure} 
    \begin{subfigure}{0.30 \linewidth}
    \includegraphics[width=\linewidth]{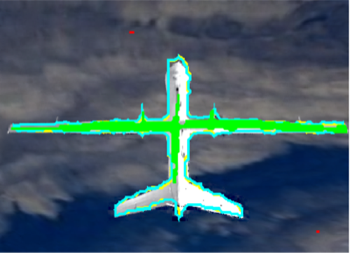} 
    \label{figinit:b} 
  \end{subfigure}

  \begin{subfigure}{0.30\linewidth}
    \includegraphics[width=\linewidth]{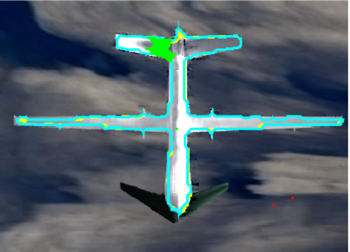} 
    \label{figinit:c} 
  \end{subfigure} 
  \begin{subfigure}{0.29\linewidth}
    \includegraphics[width=\linewidth]{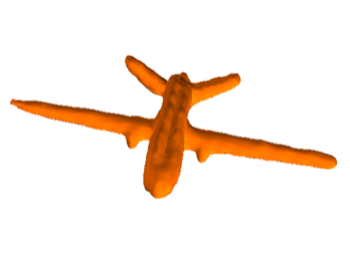} 
    \label{figinit:d} 
   \end{subfigure} 
   \caption{Example where interactions are added to the wing-tips which are darker than ambient clutter of clouds and additional shape complication due drone thinness in wings.}
  \label{fig:reaper22} 
\end{figure}

\begin{figure}[!t]
\center
\begin{subfigure}{0.30 \linewidth}
    \includegraphics[width=\linewidth]{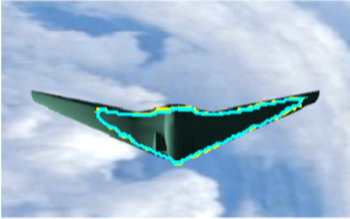} 
   \caption{} 
    \label{figinit:a}
   \end{subfigure} 
    \begin{subfigure}{0.30 \linewidth}
    \includegraphics[width=\linewidth]{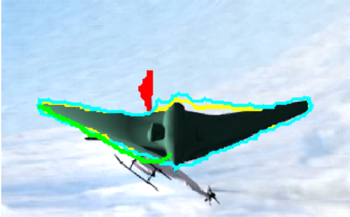} 
  \caption{} 
    \label{figinit:b} 
  \end{subfigure}

  \begin{subfigure}{0.30\linewidth}
    \includegraphics[width=\linewidth]{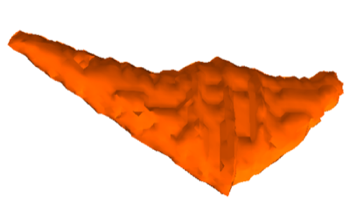} 
 \caption{} 
    \label{figinit:c} 
  \end{subfigure}
  \begin{subfigure}{0.29\linewidth}
    \includegraphics[width=\linewidth]{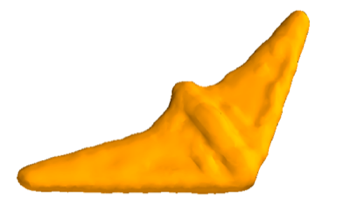} 
 \caption{}  
    \label{figinit:d} 
   \end{subfigure} 
   \caption{Complications due to varied illumination conditions where interactions are added to the left wing-tip.}
  \label{fig:sentinel} 
\end{figure}

More importantly, irrespective of the underlying model chosen for reconstruction, there will exist assumptions that are violated possibly due variety of image artifacts such as \textbf{noise, clutter, and/or model assumptions itself}.  That is, for the chosen reconstruction autonomous model, we make the classic assumption that the scene is ``mean-separable'' and piecewise constant.  Of course, while there exists other more advances models, such a model helps illustrate where operator feedback may override basic fallible assumptions.  Figure \ref{fig:real_cup} presents a scene in which such piecewise assumption is violated along minor camera miscalibrations.  Additional scenes for which such assumption is violated can be seen in Figure \ref{fig:reaper22} which aims to reconstructs a predator drone in a seemingly distinguishable background of clouds yet fails without operator input.  In the context of stereoscopic reconstruction, overcoming non-uniform illumination is yet another tacit challenge.  Figure \ref{fig:sentinel} presents a scene where reconstruction of a sentinel drone fails due to tacit illumination on the ailerons that varies over the dataset. This is in part, due to illumination on the left wing which is consequently lower than the right wing.  Utilizing operator input, the reconstruction results are demonstrated.

\begin{figure*}[!t]
\centering
\begin{subfigure}{\linewidth}
\centering
    \begin{subfigure}{.15 \linewidth}
    \centering
    \includegraphics[width=\textwidth]{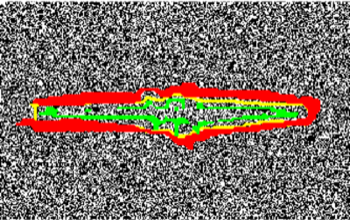}
  \end{subfigure}
  \begin{subfigure}{.15 \linewidth}
\centering
  \includegraphics[width=\textwidth]{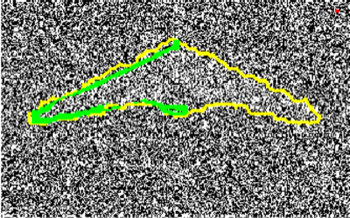}%
  \end{subfigure} 
    \begin{subfigure}{.15\linewidth}
\centering
    \includegraphics[width=\textwidth]{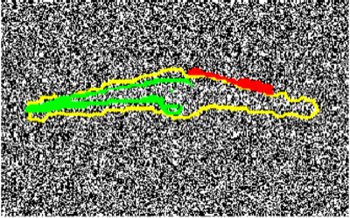}%
   \end{subfigure} 
  \begin{subfigure}{.15 \linewidth}
 \centering
    \includegraphics[width=\textwidth]{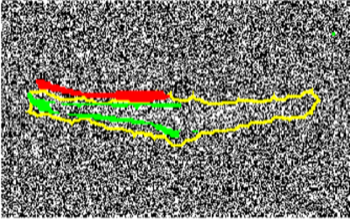}%
   \end{subfigure} 
     \begin{subfigure}{.17\linewidth}
\centering
\includegraphics[width=\textwidth]{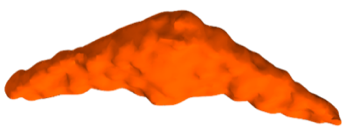}%
   \end{subfigure} 
\end{subfigure}
\\
\begin{subfigure}{\linewidth}
\centering
       \begin{subfigure}{.15 \linewidth}
    \centering
    \includegraphics[width=\textwidth]{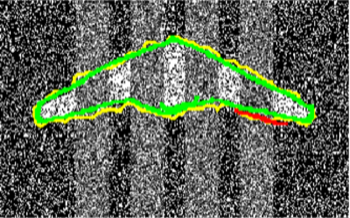}
  \end{subfigure}
  \begin{subfigure}{.15 \linewidth}
\centering
  \includegraphics[width=\textwidth]{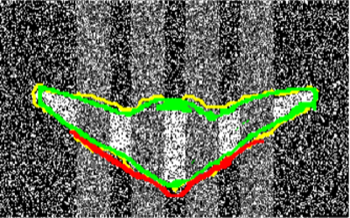}%
  \end{subfigure} 
    \begin{subfigure}{.15\linewidth}
\centering
    \includegraphics[width=\textwidth]{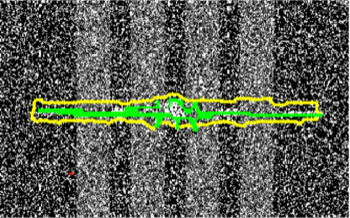}%
   \end{subfigure} 
  \begin{subfigure}{.15 \linewidth}
 \centering
    \includegraphics[width=\textwidth]{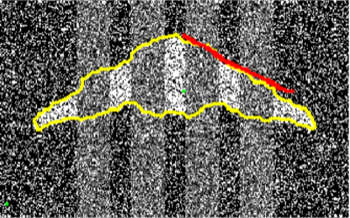}%
   \end{subfigure} 
     \begin{subfigure}{.17\linewidth}
\centering
\includegraphics[width=\textwidth]{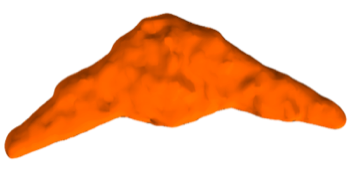}%
   \end{subfigure} 
   \end{subfigure}
\caption{Visual Synthetic Sentinel Drone Reconstruction Under Noise Conditions Corresponding to Table I.  Top Row:  Several Views Showing 90\% Noise.   Bottom Row: Several Views Showing 50\% Noise with 37\% Occlusion Over Image Domain.}
\label{fig:noiseVisual}
\end{figure*}

\begin{figure}[!t]
\centering
    \includegraphics[width=\linewidth]{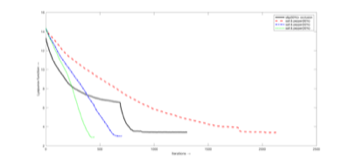} 
   \caption{Lyapunov function decay plots of noise scenes.  Black: 50\% Noise with Occlusion.  Red: Noise at 90\%. Blue: Noise at 50\%. Green: Noise at 30\%.  The ``knee-like'' decrease in error in black/red signals are due to user-input.}
  \label{fig:decayRates} 
\end{figure}

To further the idea in a \textbf{quantitive non-subjective manner}, we conduct numerical noise experiments on reconstruction of a synthetic scene of a sentinel drone which can be seen in Figure \ref{fig:noiseVisual} and Table \ref{tab:comp-table}.  Ultimately, if the operator requires intensive work to assist the autonomous counterpart in such situations, then manual operator would suffice (or desire for improved built autonomy).  This said, Table \ref{tab:comp-table} presents \textbf{efficiency} results as the amount of user input is needed (in terms of \% ``actions'' per view, \% relabeling of pixels) compared to increased output (in terms of true and false positive rate pixel labels).  For example, the second row can be stated that under 30\% noise with only one action (user-input) on 95\% of the views which amounts to only 2.7\% pixel relabeling per image view, the true positive rate increases from 78.4\% to 99.2\%.  This is repeated on several versions of noise and occlusion, two of which are seen from different views in Figure \ref{fig:noiseVisual}.  \emph{Nevertheless, the key application point of view here is that such failures of such reconstruction methods due to imaging artifacts such as noise can be naturally recover with minimal effort with human in-loop collaboration.} In addition to such results, we provide corresponding Lyapunov decay rates to such scenes in Figure \ref{fig:decayRates}.

Lastly, we note significant work on methods that use ``feature''-based methods that rely on correspondences combined with machine learning to perform reconstruction tasks \cite{dai2017scannet,chen2015deep,gwak2017savarse,zhao2018good}.  While the thematic aspect of this paper is not discuss the rigors of such methods compared to the proposed underlying autonomous method, it is worthwhile to note that under such noisy situations, such correspondence methods (dependent on structural image information) began to suffer.  Here, the geometric method proposed can be considered a ``coarse'' approach to tackle such ``featureless'' environments.  This said, future work will focus on fusing such correspondence-based and learning approaches in hopes to define a notion of image integrity and leverage recent learning success on data that is indeed well-structured.
  
\section{CONCLUSIONS AND FUTURE WORKS}
In this paper, we have proposed a feedback control framework to guide the dynamics of an evolving surface in the context of multi-view stereoscopic reconstruction.  This is done to ensure robustness in presence of low-fidelity datasets. From an optimization standpoint, the reconstruction minima which we often seek (due to modeling imperfections) may not coincide with user expectations.  As opposed to defining complex models for which overfitting may arise, we incorporate a user-defined input in-loop and ``on-the-fly'' from a feedback control perspective.  We show the resulting framework is stable via Lyapunov analysis and from a practical standpoint, there is an increase in efficiency through a human-autonomous collaboration in shape reconstruction.  Mathematically, the thematic interest is the interplay of geometry and control, namely how notions of curvature from geometry infer convergence and for this note, a notion of autonomous trust to user-input. This said, future work will entail a much closer analysis in regards to how Gaussian curvature infers convergence as well as the study of a problem in a distributed optimization sense, non-constant and time-delayed inputs as well as the inclusion of stochastic optimal control to further characterize operator uncertainty.    
\begin{table}[h]
\begin{tabular}{|c|c|c|c|c|}
\hline
Noise & Interactions & \% Pixels & True & False \\  & \% View &  & Pos. Rate & Pos. Rate   \\\hline
30\% & (0, 0) & 0\% & 78.4\% & 0.96\% \\ \hline
30\% & (1, 95\%) & 2.7\% & 99.2\% & 3.09\%  \\ \hline
50\% & (0, 0) & 0\% & 47.6\% & 0.2\%  \\ \hline
50\% & (1, 95\%) & 2.7\% & 99.8\% & 4.8\% \\ \hline
90\% & (0, 0) & 0\% & 21.8\% & 1\% \\ \hline
90\% & (1, 95\%) & 2.7\% & 99.9\% & 14.3\%  \\ \hline
90\% & \begin{tabular}[c]{@{}c@{}}(1, 36\%)\\ (2, 60\%)\end{tabular} & \begin{tabular}[c]{@{}c@{}}6.3\%\\ (2.7+3.6)\end{tabular} & 99.9\% & 4.92\%  \\ \hline
\begin{tabular}[c]{@{}l@{}}noise: 50\%+\\ occlusion: 37\%\end{tabular} & \begin{tabular}[c]{@{}c@{}}(1, 36\%)\\ (2, 60\%)\end{tabular} & \begin{tabular}[c]{@{}c@{}}6.7\%\\ (2.7+4)\end{tabular} & 99.6\% & 4.3\% \\ \hline
\end{tabular}%
\caption{Comparative analysis with noise and occlusion for the synthetic example of the Sentinel drone.}
\label{tab:comp-table}
\end{table}

\begin{figure*}[!t]
\centering

    \begin{subfigure}{0.14 \linewidth}
    \centering
    \includegraphics[width=\textwidth]{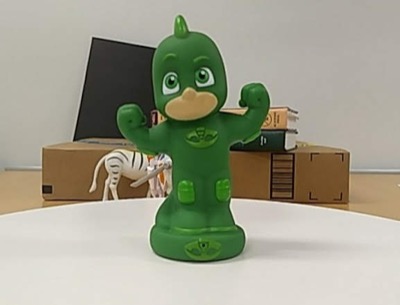}
  \end{subfigure}
  \begin{subfigure}{0.14\linewidth}
\centering
  \includegraphics[width=\textwidth]{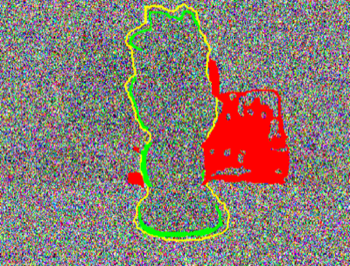}%
  \end{subfigure} 
    \begin{subfigure}{0.14\linewidth}
\centering
    \includegraphics[width=\textwidth]{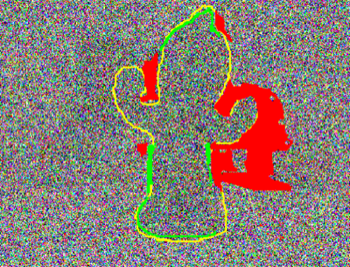}%
   \end{subfigure} 
  \begin{subfigure}{0.14\linewidth}
 \centering
    \includegraphics[width=\textwidth]{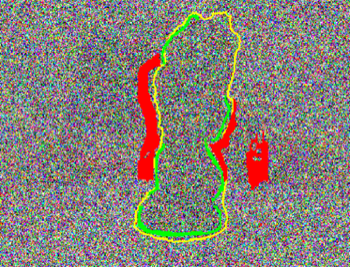}%
   \end{subfigure} 
     \begin{subfigure}{0.14\linewidth}
\centering
\includegraphics[width=\textwidth]{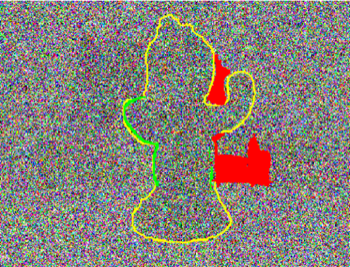}%
   \end{subfigure} 
    \begin{subfigure}{0.14\linewidth}
 \centering
\includegraphics[width=.55\textwidth]{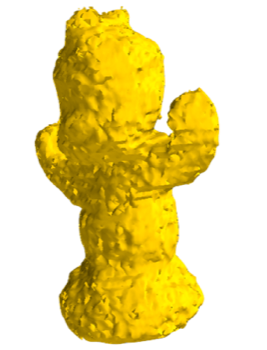}
  \end{subfigure}

    \begin{subfigure}{0.14 \linewidth}
    \centering
    \includegraphics[width=\textwidth]{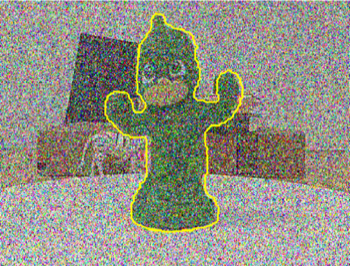}
  \end{subfigure}
  \begin{subfigure}{0.14\linewidth}
\centering
  \includegraphics[width=\textwidth]{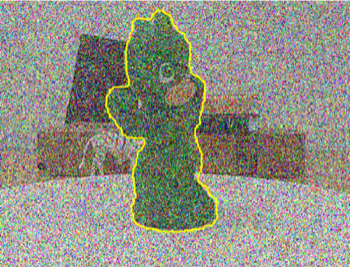}%
  \end{subfigure} 
    \begin{subfigure}{0.14\linewidth}
\centering
    \includegraphics[width=\textwidth]{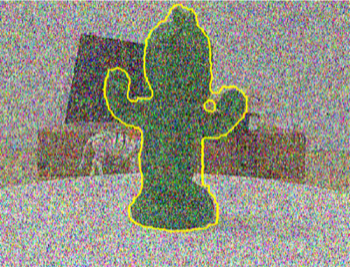}%
   \end{subfigure} 
  \begin{subfigure}{0.14\linewidth}
 \centering
    \includegraphics[width=\textwidth]{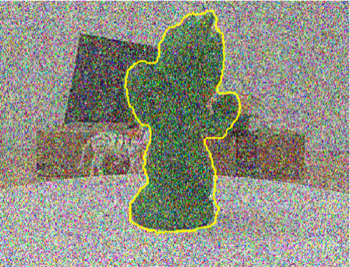}%
   \end{subfigure} 
     \begin{subfigure}{0.14\linewidth}
\centering
\includegraphics[width=\textwidth]{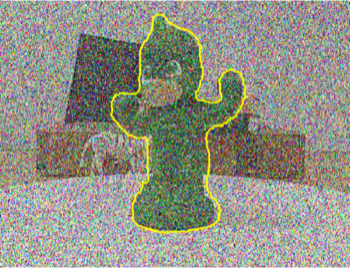}%
   \end{subfigure} 
    \begin{subfigure}{0.14\linewidth}
 \centering
\includegraphics[width=.8\textwidth]{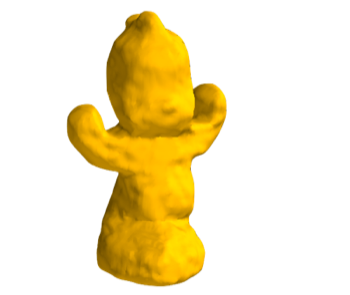}
  \end{subfigure}%

    \begin{subfigure}{0.14 \linewidth}
    \centering
    \includegraphics[width=\textwidth]{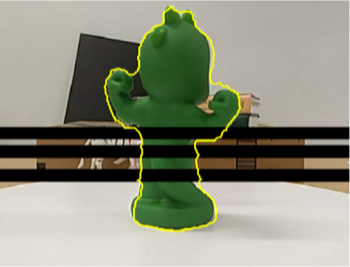}
  \end{subfigure}
  \begin{subfigure}{0.14\linewidth}
\centering
  \includegraphics[width=\textwidth]{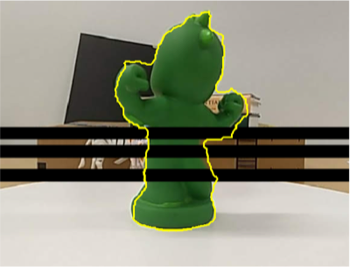}%
  \end{subfigure} 
    \begin{subfigure}{0.14\linewidth}
\centering
    \includegraphics[width=\textwidth]{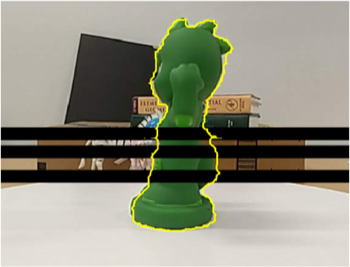}%
   \end{subfigure} 
  \begin{subfigure}{0.14\linewidth}
 \centering
    \includegraphics[width=\textwidth]{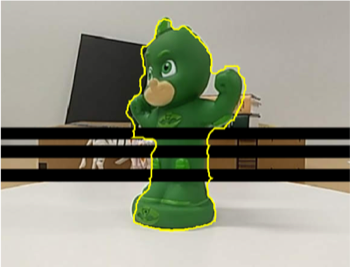}%
   \end{subfigure} 
     \begin{subfigure}{0.14\linewidth}
\centering
\includegraphics[width=\textwidth]{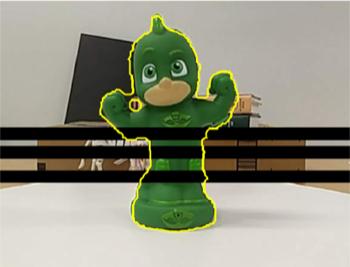}%
   \end{subfigure} 
    \begin{subfigure}{0.14\linewidth}
 \centering
\includegraphics[width=.8\textwidth]{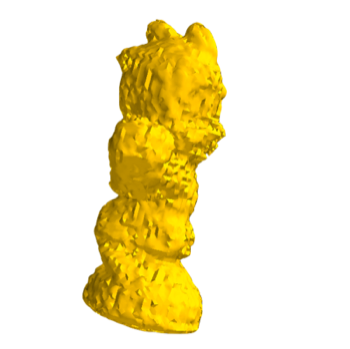}
  \end{subfigure}
  
   \caption{3D Reconstruction of a Toy Figurine.  Top Row, First Image:  Original Image of One View.  Top Row:  Significant Noise Applied to All Views with 3D Reconstruction.  Middle Row:  Moderate Noise Applied to All View with 3D Reconstruction.  Bottom Row:  Induced Camera Artifacts with 3D Reconstruction.  Note:  The underlying algorithm assumes object and background are ``mean-separable'' (e.g., such scenes are difficult corresponding to underlying autonomous model).}
  \label{fig:real_cup} 
\end{figure*}

\footnotesize{
\bibliographystyle{IEEEtran}

}

\end{document}